\def\year{2022}\relax
\documentclass[letterpaper]{article} 
\usepackage{aaai22}  
\usepackage{times}  
\usepackage{helvet}  
\usepackage{courier}  
\usepackage[hyphens]{url}  
\usepackage{graphicx} 
\usepackage{natbib}  
\usepackage{caption} 
\DeclareCaptionStyle{ruled}{labelfont=normalfont,labelsep=colon,strut=off} 
\usepackage{algorithm}
\usepackage{algorithmic}

\usepackage{newfloat}
\usepackage{listings}
\floatstyle{ruled}
\newfloat{listing}{tb}{lst}{}
\floatname{listing}{Listing}
\usepackage[utf8]{inputenc}
\usepackage[T1]{fontenc}
\usepackage{hyphenat}
\usepackage{xspace}
\usepackage{amsmath}
\usepackage{amsfonts}
\usepackage{url}
\usepackage{booktabs}
\usepackage{multirow}
\usepackage{subfigure}
\usepackage{makecell}
\usepackage{caption}
\usepackage{minibox}
\usepackage{bbm}
\usepackage{mathtools}
\usepackage{marvosym}
\usepackage{ifthen}
\usepackage{textcomp}
\usepackage{enumitem}
\usepackage{verbatim}
\usepackage{algorithm}
\usepackage{algorithmic}
\usepackage{numprint}
\usepackage{balance}
\usepackage{xcolor}

\usepackage{amsthm}
\theoremstyle{plain}

\newcommand{\abbrevStyle}[1]{#1}

\newcommand{\ie}{\abbrevStyle{i.e.}\xspace}
\newcommand{\eg}{\abbrevStyle{e.g.}\xspace}
\newcommand{\cf}{\abbrevStyle{cf.}\xspace}

\newcommand{\Eqnref}[1]{Eq.~\ref{#1}}

\newcommand{\Tabref}[1]{Table~\ref{#1}}
\newcommand{\Figref}[1]{Fig.~\ref{#1}}

\newcommand{\xhdr}[1]{\vspace{1.5mm}\noindent{{\bf #1.}}}

\newcommand{\xhdrNoPeriod}[1]{\vspace{1.7mm}\noindent{{\bf #1}}}

\newcommand{\textcite}[1]{\citeauthor{#1} \shortcite{#1}}

\newcommand{\hide}[1]{}

\newcommand{\emoji}[1]{\includegraphics[height=1.10\fontcharht\font`\A]{emoji_images/#1.png}}

\makeatletter
\newcommand{\iffont}[2]{\ifthenelse{\equal{\f@family}{#1}}{#2}{}}
\makeatother

\iffont{ptm}{
  \usepackage{mathptmx}

  \DeclareSymbolFont{greek}{OML}{cmm}{m}{n}
  \DeclareMathSymbol{\alpha}{\mathalpha}{greek}{"0B}
  \DeclareMathSymbol{\beta}{\mathalpha}{greek}{"0C}
  \DeclareMathSymbol{\gamma}{\mathalpha}{greek}{"0D}
  \DeclareMathSymbol{\delta}{\mathalpha}{greek}{"0E}
  \DeclareMathSymbol{\epsilon}{\mathalpha}{greek}{"0F}
  \DeclareMathSymbol{\zeta}{\mathalpha}{greek}{"10}
  \DeclareMathSymbol{\eta}{\mathalpha}{greek}{"11}
  \DeclareMathSymbol{\theta}{\mathalpha}{greek}{"12}
  \DeclareMathSymbol{\iota}{\mathalpha}{greek}{"13}
  \DeclareMathSymbol{\kappa}{\mathalpha}{greek}{"14}
  \DeclareMathSymbol{\lambda}{\mathalpha}{greek}{"15}
  \DeclareMathSymbol{\mu}{\mathalpha}{greek}{"16}
  \DeclareMathSymbol{\nu}{\mathalpha}{greek}{"17}
  \DeclareMathSymbol{\xi}{\mathalpha}{greek}{"18}
  \DeclareMathSymbol{\pi}{\mathalpha}{greek}{"19}
  \DeclareMathSymbol{\rho}{\mathalpha}{greek}{"1A}
  \DeclareMathSymbol{\sigma}{\mathalpha}{greek}{"1B}
  \DeclareMathSymbol{\tau}{\mathalpha}{greek}{"1C}
  \DeclareMathSymbol{\upsilon}{\mathalpha}{greek}{"1D}
  \DeclareMathSymbol{\phi}{\mathalpha}{greek}{"1E}
  \DeclareMathSymbol{\chi}{\mathalpha}{greek}{"1F}
  \DeclareMathSymbol{\psi}{\mathalpha}{greek}{"20}
  \DeclareMathSymbol{\omega}{\mathalpha}{greek}{"21}
  \DeclareMathSymbol{\varepsilon}{\mathalpha}{greek}{"22}
  \DeclareMathSymbol{\vartheta}{\mathalpha}{greek}{"23}
  \DeclareMathSymbol{\varpi}{\mathalpha}{greek}{"24}
  \DeclareMathSymbol{\varrho}{\mathalpha}{greek}{"25}
  \DeclareMathSymbol{\varsigma}{\mathalpha}{greek}{"26}
  \DeclareMathSymbol{\varphi}{\mathalpha}{greek}{"27}
  \DeclareSymbolFont{otone}{OT1}{cmr}{m}{n}
  \DeclareMathSymbol{\Gamma}{\mathalpha}{otone}{0}
  \DeclareMathSymbol{\Delta}{\mathalpha}{otone}{1}
  \DeclareMathSymbol{\Theta}{\mathalpha}{otone}{2}
  \DeclareMathSymbol{\Lambda}{\mathalpha}{otone}{3}
  \DeclareMathSymbol{\Xi}{\mathalpha}{otone}{4}
  \DeclareMathSymbol{\Pi}{\mathalpha}{otone}{5}
  \DeclareMathSymbol{\Sigma}{\mathalpha}{otone}{6}
  \DeclareMathSymbol{\Upsilon}{\mathalpha}{otone}{7}
  \DeclareMathSymbol{\Phi}{\mathalpha}{otone}{8}
  \DeclareMathSymbol{\Psi}{\mathalpha}{otone}{9}
  \DeclareMathSymbol{\Omega}{\mathalpha}{otone}{10}
  \DeclareSymbolFont{syms}{OML}{cmm}{m}{it}
  \DeclareMathSymbol{\partial}{\mathord}{syms}{"40}
  \DeclareMathAlphabet{\mathbold}{OML}{cmm}{b}{it}
  \DeclareSymbolFont{largesymbols}{OMX}{cmex}{m}{n}

}

\title{On the Context-Free Ambiguity of Emoji}

\author{
Justyna Częstochowska,\thanks{These authors contributed equally.}\textsuperscript{\rm  \emoji{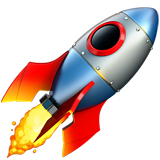}}
Kristina Gligori\'c,$^*$\textsuperscript{\rm  \emoji{rocket}}
Maxime Peyrard,\textsuperscript{\rm \emoji{rocket}}
Yann Mentha,\textsuperscript{\rm  \emoji{rocket}}
Michał Bień,\textsuperscript{\rm  \emoji{rocket}}
Andrea Grütter,\textsuperscript{\rm \emoji{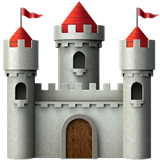}}
Anita Auer,\textsuperscript{\rm \emoji{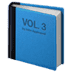}}
Aris Xanthos,\textsuperscript{\rm \emoji{unil}}
Robert West\textsuperscript{\rm  \emoji{rocket}}
}
\affiliations{

\textsuperscript{\rm \emoji{rocket}}EPFL, 
\textsuperscript{\rm \emoji{castle}}University of Zurich,
\textsuperscript{\rm \emoji{unil}}University of Lausanne\\
\{justyna.czestochowska, kristina.gligoric, maxime.peyrard, yann.mentha, michal.bien, robert.west\}@epfl.ch, \\
andrea.gruetter@es.uzh.ch, \{anita.auer, aris.xanthos\}@unil.ch
}

\begin{document}

\widowpenalty=10000
\clubpenalty=10000

\maketitle

\begin{abstract}
Due to their pictographic nature, emojis come with baked-in, grounded semantics. Although this makes emojis promising candidates for new forms of more accessible communication, it is still unknown to what degree humans agree on the inherent meaning of emojis when encountering them outside of concrete textual contexts. 
To bridge this gap, we collected a crowdsourced dataset (made publicly available) of one-word descriptions for 1,289 emojis presented to participants with no surrounding text. The emojis and their interpretations were then examined for ambiguity.
We find that, with 30 annotations per emoji, 16 emojis (1.2\%) are completely unambiguous, whereas 55 emojis (4.3\%) are so ambiguous that the variation in their descriptions is as high as that in randomly chosen descriptions. Most emojis lie between these two extremes.
Furthermore, investigating the ambiguity of different types of emojis, we find that emojis representing symbols from established, yet not cross\hyp culturally familiar code books (e.g., zodiac signs, Chinese characters) are most ambiguous.
We conclude by discussing design implications.
\end{abstract}

\section{Introduction}
\label{sec:intro}


For over a decade, emojis have been playing an increasingly important role in online communication. As of September 2021, there are 3,633 emojis in the Unicode standard \cite{Unicode2021EmojiV14.0}, and the number is growing, providing users with more ways to express increasingly complicated concepts. 
Consequently, emojis have received much attention from researchers. Various fields, including natural language processing, human\hyp computer interaction, and Web and social media research, study the usage and function of emojis.

Beyond today's prevalent use cases (social media and instant messaging), emojis have untapped potential for facilitating communication in other contexts as well. While letters, syllables, and words are arbitrary and highly abstract constructs that require a long time to master, emojis come already packed with richly grounded semantics. Emojis can thus be leveraged, \eg, in learning and education \cite{GillesDoiron2018EMOJIS:EDUCATION} or to describe complex ideas to broad audiences \cite{AndyThomason2014FinallyUnderstand}.





However, it is unknown which emojis can be used for such goals. As a first step, it is necessary to establish how much people agree about the context-free interpretation of individual emojis. Doing so has broad implications for social media and Web research, communication studies, education, and more. Beyond research communities, identifying which emojis are ambiguous is helpful to online communities and emoji designers to prevent introducing emojis with a high potential for miscommunication. Additionally, studying context-free emoji semantics informs us which concepts can or cannot be easily communicated with emojis.



Despite the practical importance of these questions, it is difficult to approach them with available datasets. Social media content carries inherent selection biases, and emoji studies leveraging social media have been questioned for their generalizability \cite{Herring2020GenderFunctions}. Additionally, as emojis are almost only used in context, it is difficult to infer context-free interpretations. To complicate matters further, the meaning of emojis on social media evolves with time \cite{AlexanderRobertson2021Semantic20122018}, making it difficult to study their intrinsic semantics.



Here we ask: \emph{Do individuals interpret emojis similarly? Which emojis have the potential to be used in future communication scenarios, and to what extent?} Previous work on the ambiguity of emojis focused on differences between platforms \cite{Shurick2020WhatsVendors} and their usage in context \cite{Miller2017UnderstandingMiscommunication}. Most closely related past studies focused on frequently used and anthropomorphic emojis \cite{Shurick2020WhatsVendors,Miller2017UnderstandingMiscommunication, Miller2016BlissfullyEmoji}, discarding many available emojis. Whereas a lot is known about emoji sentiment and usage in context \cite{Novak2015SentimentEmojis}, less is known about emoji semantics beyond sentiment and context-free emoji interpretation. To bridge this gap, we designed and executed a crowdsourced study examining an exhaustive set of emojis, many of which are rarely used in online communication. We studied their interpretation in the absence of any textual context. 
Using the resulting novel dataset of emoji annotations, we address the following research questions:

\noindent \textbf{RQ1:} To what degree do people agree when interpreting emojis?

\noindent \textbf{RQ2:} What types of emojis are most and least ambiguous?


\section{Related Work}
\label{sec:related}

\xhdr{Emojis: interpretation and meaning} Previous research has shown that emojis are often misunderstood \cite{Miller2016BlissfullyEmoji,Miller2017UnderstandingMiscommunication}.
Misunderstanding is sometimes related to how the emoji's design is interpreted in context or the way it is shown on the receiving side.
In particular, in 2016, \citeauthor{Miller2016BlissfullyEmoji} examined interpretations of the 25 most popular anthropomorphic emojis without context, across five popular platforms. The study compared differences in sentiment and semantics to identify the most ambiguous emojis.
In 2017, \citeauthor{Miller2017UnderstandingMiscommunication} conducted a similar study comparing sentiment variability with and without context, for 10 anthropomorphic emojis.
An extensive dataset of emoji senses was created by \citeauthor{Wijeratne2017EmojiNet:Discovery}, linking Unicode emoji representations to their meanings automatically extracted from the Web. 
Recent studies of emoji meaning provide a longitudinal perspective \cite{Barbieri2018ExploringLens,AlexanderRobertson2021Semantic20122018}. 
Our work studies the intrinsic ability of emojis to convey information, independent of the textual context they are used in. In contrast to \citet{Miller2016BlissfullyEmoji, Miller2017UnderstandingMiscommunication}, who focused on small subsets of anthropomorphic emojis, we consider a far more exhaustive set of emojis (see \Tabref{table:representatives}).

\xhdr{The aspiration of an emoji\hyp based language} There is growing interest in the linguistic purposes of emojis \cite{Naaman2017MojiSem:Context} and their potential to emerge as a graphical language \cite{Ge2018CommunicativeWeibo}. There have been multiple informal initiatives to create an emoji language, such as the attempt to translate \textit{Moby Dick} into a sequence of emojis.%
\footnote{https://www.kickstarter.com/projects/fred/emoji-dick}
Such efforts demonstrate the potential for viewing emojis as the atomic units of graphical and intuitive language that could remove accessibility barriers inherent in standard written natural languages. 



\xhdr{Emojis, social media, and natural language processing} Social media researchers have been examining the ways social media users use, interpret, and express emotions and information through emojis. It is well known that emojis shape online language \cite{Feldman2021EmojisCommunication,Pavalanathan2016ViewMonday}. Emoji usage can also be a proxy for studying human behavior; \eg, emojis are a powerful indicator in the context of crisis events \cite{Santhanam2018IEvents} and can be used to identify group belonging \cite{Jones2021TheTwitter}. Researchers have also been analyzing the use of gender and skin-tone modifiers \cite{Barbieri2018HowTwitter, Robertson2020EmojiModifiers, Robertson2021IdentityTwitter}. As emojis became a standard element of online language, a need to computationally process them emerged. Creating meaningful, latent emoji representations \cite{ Eisner2016Emoji2vec:Description} and emoji prediction tasks
\cite{Barbieri2018InterpretableLSTMs} gained importance in NLP. We thus note that our annotations can be used to compute or augment emoji representations and thus support the social media and NLP research communities.



\section{Methods and Data}
\label{sec:methods}

\xhdr{Emoji selection} We selected emojis for our study as follows. Starting from all available 3,633 emojis, we removed letters, numbers, flags, and gendered and skin-toned anthropomorphic emojis, considering only neutral versions of such emojis. We also removed variations of the same emoji (\eg, family with three or four members). This resulted in the final set of 1,289 emojis. Furthermore, we collected emoji categories from Emojipedia.org and hand\hyp crafted a categorization extending the seven existing categories to 20 fine-grained types, outlined in \Tabref{table:representatives}.

\xhdr{Annotation process} We collected emoji interpretations using the Amazon Mechanical Turk (AMT) crowdsourcing platform. Each participant was asked to ``Describe emojis with a single, accurate word''. Each task consisted of 10 emojis. All participants had to be at least 18 years old, speak English, reside in the USA, have a 99\% approval rate, and have completed at least 500 tasks on AMT before. Our annotator compensation was in line with ethical guidelines for AMT \cite{Whiting2019FairCode}. Each emoji was annotated by 30 unique participants. This number was chosen via pilot studies (150 annotations for 12 emojis), showing that, as the number of annotations increases beyond 30, the word distribution remained robust. Overall, we collected 38,670 annotations, for an average of 82.5 annotations per participants. In total, there were 445 unique participants. We asked participants to provide their age, gender, and mother tongue. The majority of annotators were native English speakers (97\%). Participants' gender was well balanced (55\% female, 44\% male, 1\% other or not stated). The average age was 38.8 ($\text{SD}=12.0$).

\xhdr{Post-processing} To improve the quality of annotations, we performed three post-processing steps: low quality annotator detection, validation of honeypots, and spelling correction. We performed detection of annotators with low annotation quality by identifying those who used the same word for all emojis in a task. We discarded one annotator whose vocabulary size was less than 80\% of the number of assigned emojis. To further ensure the quality, one unquestionably non\hyp ambiguous emoji was placed in every task. Annotations whose answers did not match any words from the expected set (\eg,  ``apple'' for \emoji{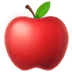}, ``pizza'' for \emoji{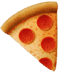}, ``carrot'' for \emoji{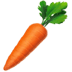}) were excluded. Finally, to account for spelling mistakes, we cross-checked word validity using the PyEnchant library.

\newcommand{\Mode}{v^*}
   
\xhdr{Measuring semantic variation} We use word embeddings (\ie, representations of words as numerical vectors) to quantify semantic similarity. To measure the extent to which annotators agree about emoji meaning, we calculate the dispersion of emoji annotations in a similar way to \citeauthor{Miller2016BlissfullyEmoji}, using GloVe vectors \cite{Pennington2014GloVe:Representation} of dimensionality 200 \cite{Rehurek2011GensimpythonModelling}. Let $V$ denote the set of distinct words used by respondents to annotate the considered emoji, which we will call the emoji’s \textit{vocabulary}; $f_v$ stands for the relative frequency of word $v$ in the emoji's annotations
and $\Mode := \arg \max_{v \in V} f_v$ is the mode annotation, i.e., the most frequent word in $V$. 
We then define the emoji’s semantic variation as the weighted average of the cosine distance between the embedding $e_v$ of each word $v \in V$ and the embedding $e_{\Mode}$ of the mode annotation in $V$:

\begin{equation}
\text{semantic variation} = \sum_{v \in V} f_v \cdot (1 - \cos(e_v, e_{\Mode}))
\label{eq:variation}
\end{equation}


\begin{table}
\centering
\tiny
\setlength\tabcolsep{2pt}
\begin{tabular}{l|l|c|c}
        \textbf{Category} & \textbf{Category description} & \textbf{Num.} &  \textbf{Examples}  \\ \hline 
        \hline
        objects        & household items, celebrations, stationery,  &  202  &         \emoji{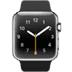} \emoji{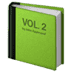} \emoji{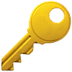} \\
        
        &  and miscellaneous objects &   &\\
        
        nature         &  scenery, animals, plants, weather &  189  &        \emoji{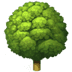} \emoji{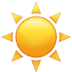} \emoji{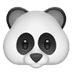}  \\
        
        travel-places  &  buildings, vehicles, landscapes  &  129 &       \emoji{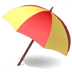} \emoji{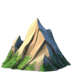} \emoji{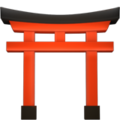}   \\ 
        
        food-drink     &  fruit, vegetables, meals, beverages and utensils   & 113  &        \emoji{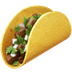} \emoji{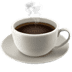} \emoji{apple}   \\
        
        faces          &  smileys and faces with their cat versions    &  111 &      \emoji{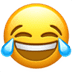} \emoji{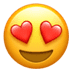} \emoji{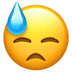}   \\
        
        people         &  anthropomorphic emojis other than professions  & 103 &     \emoji{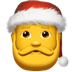} \emoji{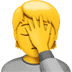} \emoji{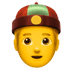}   \\
        
        activity       &  sports, music, the arts, hobbies, other activities &  78 &     \emoji{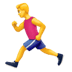} \emoji{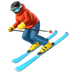} \emoji{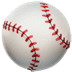}   \\
        
               &   and objects associated with them     &   &     \\

        clothes \& accessories &  umbrellas, shoes, purses, traditional clothing   &  47 &    \emoji{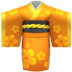} \emoji{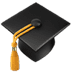} \emoji{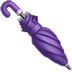}    \\
        
        symbols \& signs       &  mathematical symbols, currencies, cards suit,  & 43 &   \emoji{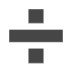} \emoji{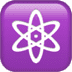} \emoji{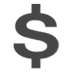}     \\
               &   punctuation marks &  &     \\

        professions            & anthropomorphic emojis representing  &  38 &  \emoji{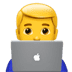} \emoji{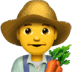} \emoji{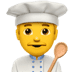}   \\
        
         & professions such as doctor and lawyer &   &   \\
        
        geometrical            & shapes of different sizes and colors such as & 34 &             \emoji{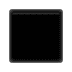} \emoji{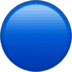} \emoji{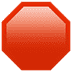}  \\
          &  large blue circle and orange diamond  &  &              \\
        
        hands \& gestures      &  gestures or activities done with hands such as  &   34 &      \emoji{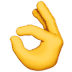} \emoji{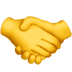} \emoji{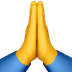}    \\
            &   folded hands and writing hands &    &        \\
        
        Japanese symbols & emojis originating in Japanese culture &  34   &      \emoji{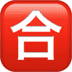} \emoji{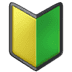} \emoji{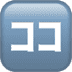}   \\
        \& objects  &  &    &       \\
        
        buttons \& mobile   &  buttons used in mobile phones  & 29 &       \emoji{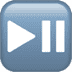} \emoji{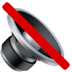} \emoji{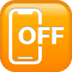}    \\
        
        public information & danger warning, priority prohibitive,   & 26 &   \emoji{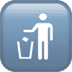} \emoji{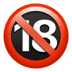} \emoji{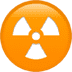}  \\
        
          symbols & facilities, or service signs & &    \\
        
        letters \& numbers      &  letters, numbers and their combinations &  21 &   \emoji{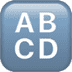} \emoji{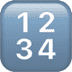} \emoji{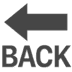}    \\
        
        hearts                  &  hearts of different colours &  19 &   \emoji{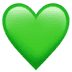} \emoji{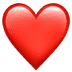} \emoji{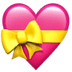}   \\
        
        arrows                  &  arrows pointing in different directions &  16 &   \emoji{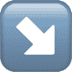} \emoji{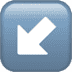} \emoji{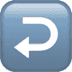}   \\
        
        astrological            & astrological zodiac signs   & 13  &     \emoji{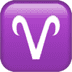} \emoji{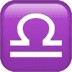} \emoji{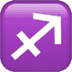}  \\
        
        religious               & symbols of major religious groups &  10 &     \emoji{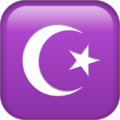} \emoji{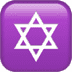} \emoji{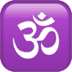}  \\
\end{tabular}
\caption{Emoji categorization. Twenty categories, category descriptions, number of emojis, and three examples.}
\label{table:representatives}
\end{table}





\section{Results}
\label{sec:results}

\begin{figure}[t]
\centering
\includegraphics[width=0.9\columnwidth]{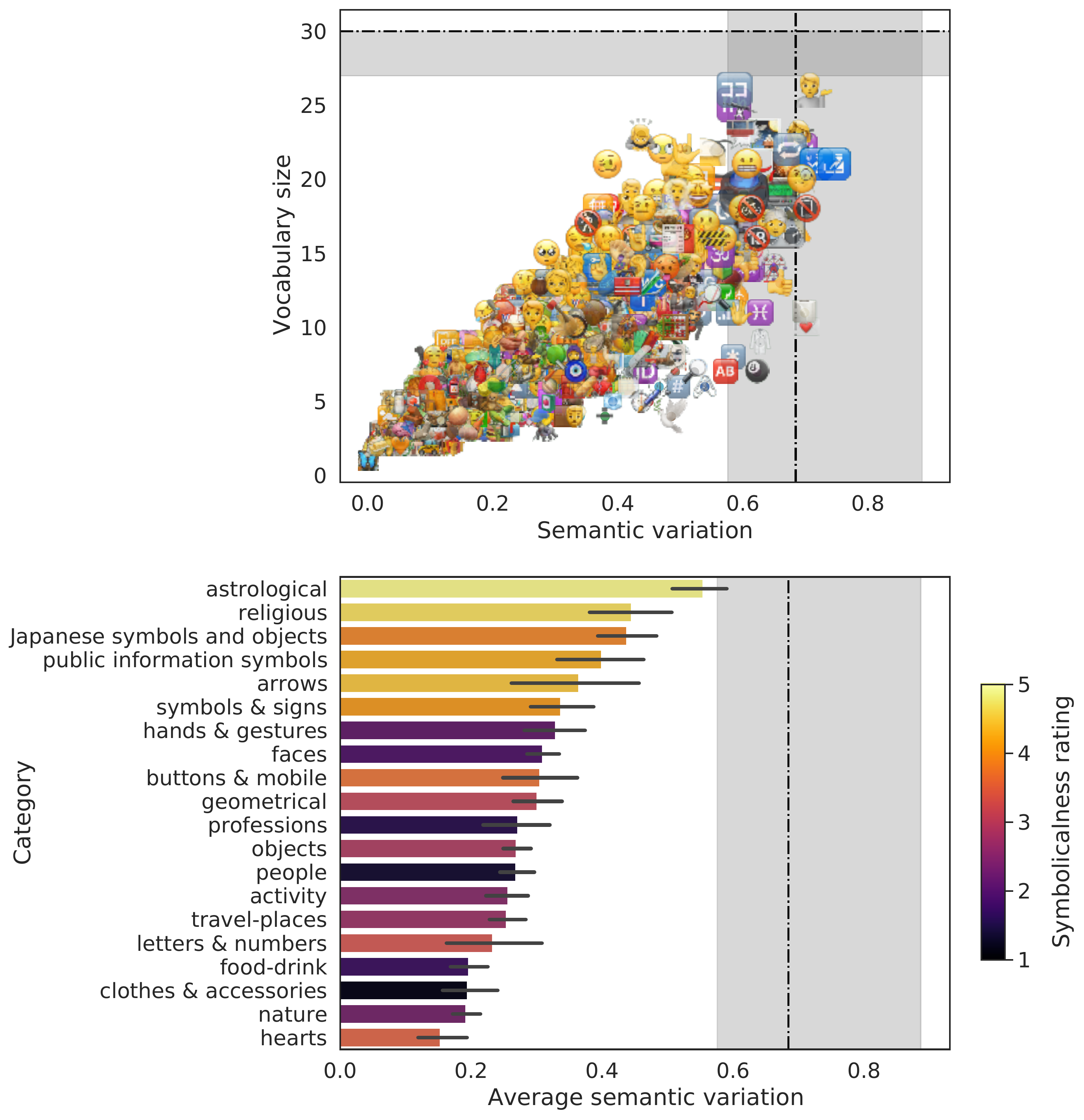}
\caption{Top: the relationship between semantic variation (x\hyp axis), and vocabulary size (y\hyp axis). Bottom:
average semantic variation across emoji categories (\cf \Tabref{table:representatives}). Color represents the extent to which emojis within a category can be seen as belonging to an established code book of symbols. Black dashed lines represent random baselines and gray bands their bootstrapped 95\% confidence intervals.}
\label{fig:figure1}
\end{figure}


\xhdrNoPeriod{RQ1: To what degree do people agree when interpreting emojis?} For each emoji, we measure the consistency among the words chosen to describe it. Consider the example of the fire emoji \emoji{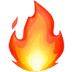}, for which one annotator used the word ``hot'' and another, ``fire''. Since the terms are different, the annotators---strictly speaking---do not agree. Yet, the words ``hot'' and ``fire'' are semantically close. We aim to capture such similarities via the notion of semantic variation (\Eqnref{eq:variation}).

To detect semantic variations significantly different from random, we compute the semantic variation of a random baseline. We sample $n=30$ random words from the distribution of vocabulary across all annotations and calculate the semantic variation. We repeat the process 1,000 times to compute 95\% confidence intervals (CI) and obtain a baseline semantic variation of $0.69$ (95\% CI $[0.57, 0.88]$).
When an emoji's semantic variation falls in the 95\% CI, its vocabulary could have equally well been a random set of words. In such cases, humans clearly do not agree in their interpretation. We repeat the same procedure to compute a random baseline with respect to vocabulary size (rather than variation), obtaining an average of 30 (\ie, all words are different), with 95\% CI $[27, 30]$. All emojis, with their semantic variation and vocabulary size, are presented in \Figref{fig:figure1}.

First, we find a strong positive correlation between vocabulary size and semantic variation (Spearman's $\rho= 0.84$, $p< 10^{-10}$). We note that emojis with a small vocabulary size can range from very ambiguous to not ambiguous at all, while the ones with a rich vocabulary have higher ambiguity.


Second, we find that 16 out of 1,289 emojis (1.2\%) reach a variation of 0, \ie, they were described with a single, unique word by all 30 annotators. These include \emoji{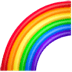} \emoji{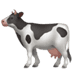} \emoji{apple} \emoji{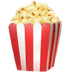} \emoji{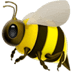} \emoji{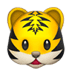} \emoji{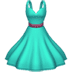} \emoji{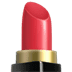} \emoji{key} \emoji{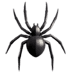} \emoji{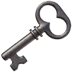} \emoji{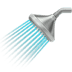} \emoji{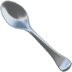} \emoji{carrot} \emoji{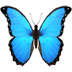} \emoji{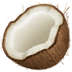}. For communication applications, these emojis are likely to be useful---they are unlikely to introduce misunderstanding.

Third, the semantic variation of 55 out of 1,289 emojis (4.3\%) falls into the random baseline confidence interval. Given the intuition that emojis come with built\hyp in semantics, it is striking that some of them exhibit such low agreement. For future communication applications, these emojis are unlikely to be useful, as they introduce high levels of ambiguity.

In summary, human agreement about the context-free meaning of emojis ranges from completely unambiguous (16 emojis) to indistinguishable from random (55 emojis), with emojis covering the whole spectrum of ambiguity. Our dataset can guide communication applications in choosing appropriate emojis to facilitate understanding.



\xhdrNoPeriod{RQ2: What types of emojis are most and least ambiguous?} We further investigate discrepancies in ambiguity, expecting different categories of emojis to exhibit different average variations. We report the average semantic variation per category in \Figref{fig:figure1}. 


Our results add nuance to the findings of \citet{Miller2016BlissfullyEmoji}, who found that anthropomorphic emojis can be more ambiguous than emojis characterizing things. In \Figref{fig:figure1}, the \textit{faces} category takes a middle place in the semantic\hyp variation ranking. Still, it is more ambiguous than objects.
Categories with the lowest average variation are \textit{food \& drink, clothes \& accessories, nature,} and \textit{hearts}.

Interestingly, the top five most ambiguous categories are the ones that emerged from further dividing the original \textit{symbols} category.
Every emoji is, of course, a symbol, but whereas some emojis derive their meaning by immediately representing the shape of commonly known objects (\eg, \emoji{apple}), those in the \textit{symbols} category refer to entries from established code books of symbols of a less immediately pictographic nature (\eg, \emoji{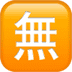} \emoji{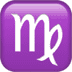}).
Such emojis are, in a sense, ``symbols of symbols'' or ``second-order symbols'', and, without prior knowledge, may be impossible to interpret.
In particular, astrological (zodiac) signs form the only category as ambiguous as the random baseline. They tended to be described with very different words or with names of other astrological signs. 
One could argue that astrological signs have an unambiguous mapping to their names, but without background knowledge, they yield ambiguous standalone interpretations.
Similarly, emojis representing Japanese signs or having origins in Japanese culture (\eg, \emoji{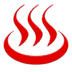} \emoji{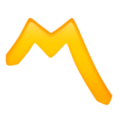})
occupy the third place in terms of semantic variation, likely due to annotators' demographics and cultural background (United States residents, native English speakers). To describe emojis of Japanese, Chinese, and Korean characters (\eg, \emoji{free-of-charge} \emoji{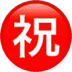} \emoji{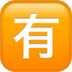}), annotators consistently used words such as: \textit{japanese, chinese, asian, sign}. This was not the case for some emojis (\eg, \emoji{beginner} \emoji{part-alternation-mark}) of whose Japanese origin annotators may not have been aware.

Based on these observations, we next investigate such ``symbol-of-symbol'' emojis in more detail.
To quantify the degree to which an emoji belongs to an established code book of symbols
(henceforth ``symbolicalness''),
two authors independently annotated all 1,289 emojis by indicating their level of agreement with the statement ``This emoji is a symbol'' on a five-point Likert scale where 1 corresponded to ``absolutely disagree'' and 5 to ``absolutely agree''.

We pre-established an annotation framework where we assigned levels from 5 to 1, respectively, to emojis representing objects and concepts that 
(5)~are established symbols and can be encountered in everyday or specialized activities (\eg,  \emoji{om-symbol} \emoji{south-east-arrow});
(4)~can have a symbolic meaning and be encountered in everyday or specialized activities (\eg,  \emoji{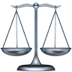} \emoji{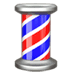} \emoji{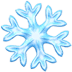});
(3)~may or may not have a symbolic meaning (\eg, \emoji{blue-circle} \emoji{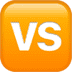});
(2)~typically do not have a symbolic meaning (\eg, \emoji{sun} \emoji{muted-speaker} \emoji{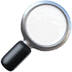});
(1)~are not established symbols (faces, gestures, people) (\eg, \emoji{ok-hand-sign} \emoji{person-facepalming} \emoji{runner}).

Following this framework, we obtained Kendall's $\tau=0.8$ ($p=1.55\times 10^{-216}$) between the authors. We computed the average symbolicalness for each emoji and averaged the values for emojis within a category to obtain the category's symbolicalness rating. We represent the rating with a color scale in \Figref{fig:figure1}. There is a weak positive correlation (Spearman's $\rho=0.25$, $p=1.61\times 10^{-19}$) between semantic variation and symbolicalness. In addition, the most ambiguous categories of emojis indeed are the ones with the highest symbolicalness rating. Even though symbols are designed to facilitate communication, our results indicate that symbolic emojis can, maybe unintuitively, be ambiguous, as their interpretation requires specific prior knowledge. 

In summary, we find that human agreement about the interpretation of emojis varies across different emoji types. The symbolicalness, or the extent to which an emoji is a symbol from an established code book of symbols, is an important dimension explaining the differences.

\section{Discussion}
\label{sec:discussion}



\xhdr{Summary of main findings} Investigating whether people interpret emojis in the same way (\textbf{RQ1}), we find that emojis come with very different amounts of prepacked semantics. Some emojis are completely unambiguous, with all annotators describing them with the same word. On the opposite, others are as ambiguous as if their descriptions were drawn at random. To support the goal of using emojis to facilitate communication, the unambiguous emojis are the best candidates to bring direct benefits. Investigating what types of emojis are ambiguous (\textbf{RQ2}), we find that different types of emojis have very different levels of agreement in interpretation. An important dimension explaining the agreement differences is the degree to which an emoji belongs to an established code book of symbols. Emojis referring to symbols require background knowledge for interpretation and are less likely to be unambiguously recognized.

Concrete objects and things can easily be illustrated by an emoji, whereas abstract ideas and concepts are harder to represent without referring to symbolic ideas from shared cultural knowledge. 
Yet to support complex communication goals, it is necessary to refer to abstract ideas. This has to be via pre-established symbols, as there is no immediately pictographic way to represent abstract concepts such as peace (\emoji{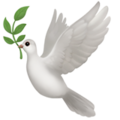} \emoji{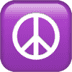}) or resistance (\emoji{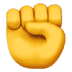}). Luckily, there exist ubiquitous symbols whose interpretations are widely agreed upon (\eg, \emoji{red-heart} for love). These symbols can be leveraged to convey complex ideas universally. 

\xhdr{Design implications} We highlight two mechanisms likely fueling measured variation in ambiguity and discuss their implications. First, the fact that concrete objects can more easily be universally described with emojis highlights the importance of considering the intended participants and their shared cultural background to appropriately choose the symbolic emojis to use. 
Second, emoji design is known to contribute to misinterpretations \cite{Miller2017UnderstandingMiscommunication}, since emojis are limited in size and need to be comprehensible even if displayed tiny; \eg, the ``pine decoration'' emoji \emoji{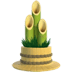} contains a fair amount of details that are difficult to display on a small scale, further jeopardizing the understanding of the concept. Our data capturing empirical emoji ambiguity can thus help make emojis more accessible and user\hyp friendly.

\xhdr{Limitations and future work} Our goal is to provide initial measurements of the ambiguity of emojis. Therefore, our study is not without its limitations. All annotators provided a single word to describe an emoji. In the future, it will be interesting to extend the study to descriptions beyond one word. Also, all annotators were English speakers residing in the United States. Future work should generalize to other cultures and languages and understand how emoji ambiguity is associated with social media usage.

\xhdr{Code and data} Code and data are publicly available at \url{https://github.com/epfl-dlab/emoji-ambiguity}.

\xhdr{Acknowledgments} This work was funded by Collaborative Research on Science and Society (CROSS), with further support from
the Swiss National Science Foundation (grant 200021\_185043), Microsoft, Google, and Facebook.

\bibliography{references.bib}

\end{document}